\documentclass[a4paper, 10pt, conference]{ieeeconf}      
\usepackage{FG2021}

\FGfinalcopy 

\IEEEoverridecommandlockouts                              

\usepackage{graphics} 
\usepackage{epsfig} 
\usepackage{mathptmx} 
\usepackage{times} 
\usepackage{amsmath} 
\usepackage{amssymb}  
\usepackage{multirow}
\usepackage{colortbl}
\usepackage{arydshln}
\usepackage{booktabs}

\def\FGPaperID{127} 

\title{\LARGE \bf
TrouSPI-Net: Spatio-temporal attention on parallel atrous convolutions and U-GRUs for skeletal pedestrian crossing prediction
}

\author{\parbox{16cm}{\centering
    {\large Joseph Gesnouin$^{1,2}$, Steve Pechberti$^{1}$, Bogdan Stanciulescu$^{2}$ and Fabien Moutarde$^{2}$}\\
    {\normalsize
    $^1$ Institut VEDECOM—Versailles, 78000 Versailles, France\\
    $^2$ Centre de Robotique, MINES ParisTech, Université PSL, 75006 Paris, France}}
}

\begin{document}

\ifFGfinal
\thispagestyle{empty}
\pagestyle{empty}
\else
\author{Anonymous FG2021 submission\\ Paper ID \FGPaperID \\}
\pagestyle{plain}
\fi
\maketitle

\begin{abstract}


Understanding the behaviors and intentions of pedestrians is still one of the main challenges for vehicle autonomy, as accurate predictions of their intentions can guarantee their safety and
driving comfort of vehicles. In this paper, we address pedestrian crossing prediction in urban traffic environments by linking the dynamics of a pedestrian’s skeleton to a binary crossing intention. We introduce TrouSPI-Net: a context-free, lightweight, multi-branch predictor. TrouSPI-Net extracts spatio-temporal features for different time resolutions by encoding pseudo-images sequences of skeletal joints’ positions and processes them with parallel attention modules and atrous convolutions. The proposed approach is then enhanced by processing features such as relative distances of skeletal joints, bounding box positions, or ego-vehicle speed with U-GRUs.
Using the newly proposed evaluation procedures for two large public naturalistic data sets for studying pedestrian behavior in traffic: JAAD and PIE, we evaluate TrouSPI-Net and analyze its performance. Experimental results show that TrouSPI-Net achieved 76\% F1 score on JAAD and 80\% F1 score on PIE, therefore outperforming current state-of-the-art while being lightweight and context-free.

\end{abstract}

\section{INTRODUCTION}

The topic of pedestrian crossing prediction has attracted significant interest in computer vision and robotics communities but remains a difficult research topic due to the great variation and complexity of its input data. Although many approaches have been proposed which report interesting results on pedestrian crossing prediction, most of the existing methods may suffer from a large model size and slow inference speed by aggregating multiple forms of perception modalities extracted by additional networks such as background context, optical flow, or pose estimation information \cite{2020arXiv200507796P,kotseruba2021benchmark,2015arXiv150308909Y,9304591,bhattacharyya2018long,7298714,Rasouli2019PedestrianAA}. 

However, in such decisive applications, a desirable action prediction model should run efficiently for real-time usage and should also be robust to a multitude of complexities and conditions. To alleviate this issue, we propose a model using only one additional network to compute poses and disregard the other perception modalities. 

Our contributions are summarized in the following:
\begin{itemize}
    \item We propose TrouSPI-Net: a scene-agnostic, lightweight, multi-branch approach that relies on pose kinematics to predict crossing behaviors. The proposed approach could be applied following the application of any additional network to compute pedestrian body poses and could be easily implemented in any embedded devices with real-time constraints since it only uses standard deep-learning operations in an euclidean grid space. 
    \item We first represent a skeleton sequence as a 2D image-like spatio-temporal continuous representation. As the scale of pedestrians' actions patterns might extend through time and is not limited by a specific temporal resolution, we extract spatio-temporal features by relying on parallel processing of 2D atrous convolutions enhanced with self-attention for multiple dilation rates. This allows TrouSPI-Net to capture features for a given pedestrian action pattern for multiple temporal resolutions. 
    \item We secondly represent a skeleton sequence as its evolution of Euclidean pairwise distances of skeletal joints over time and encode them with U-GRUs \cite{rozenberg2021asymmetrical}: a non-symmetrical bidirectional recurrent architecture designed to exploit the bidirectional temporal context and long-term temporal information for challenging skeletal dynamics having similar patterns but different outputs. This compensates for the inabilities of the first stream in learning temporal patterns invariant to locations and viewpoints. 
    \item Evaluation of TrouSPI-Net has been conducted with the freshly proposed common evaluation criteria \cite{kotseruba2021benchmark} on two standard benchmarks for pedestrian behaviors prediction: Joint Attention in Autonomous Driving (JAAD) \cite{rasouli2017ICCVW,Rasouli2017IV} and Pedestrian Intention and trajectory Estimation (PIE) \cite{rasouli2017they} public data-sets. Architecture variations and branch ablations are also presented to provide insight into our proposed multi-branch approach. 
\end{itemize}

\begin{figure}
    \centering
    \includegraphics[scale=0.6]{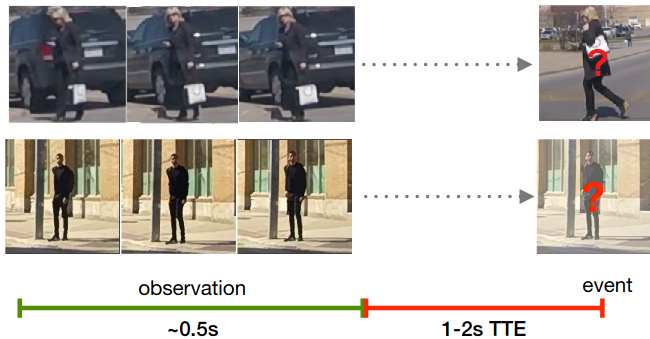}
    \caption{Pedestrian Action Prediction: the objective is to predict if the pedestrian will start crossing the street at some time $t$ given the observation of length  $m$. Figure adapted from \cite{kotseruba2021benchmark}.}
    \label{fig:presentation}
\end{figure}

\section{RELATED WORK}

Currently, the main modalities used for action recognition include RGB videos in their entirety \cite{donahue2015long,2014arXiv1412.0767T,varol2017long,Wu_2018_CVPR}, optical flow \cite{simonyan2014two,zhang2016real,sevilla2018integration,DanutPOP} and pose-based modeling \cite{devineau2018deep,hou2018spatial,2018arXiv180506184Z,yan2018spatial}, which is more detailed bellow.
\subsection{Pose-based Action Recognition}

The detection and pose estimation of humans is the first and necessary step in pose-based action recognition, of which posture analysis is an essential component. Nowadays, pose estimation approaches are not limited to use motion capture systems or depth cameras. RGB data can be used to infer 2D body poses \cite{cao2017realtime}, 3D body poses \cite{martinez2017simple} and even track people in real-time \cite{2018arXiv180200977X}. This breakthrough has stimulated the skeletal modality interest since it proved to be sufficient to describe and understand the motion of a given action without any background context.
This has made pose-based action recognition preferred over other modalities on a huge amount of real-time scenarios for human action recognition such as human-robot interaction \cite{mazhar:hal-01734739,martin2019real}, medical rehabilitative applications \cite{mousavi2014review,chang2011kinect} or pedestrian action prediction \cite{gesnouin2020predicting,fang2018pedestrian,8500657}. Some commonly used learning architectures for pose-based action recognition include 1D/2D convolutional networks \cite{devineau2018deep,pham2018learning}, recurrent networks \cite{baccouche2011sequential,shahroudy2016ntu}, a combination of one of the latter with attention mechanisms \cite{maghoumi2019deepgru,hou2018spatial} or Graph-based models \cite{2018arXiv180506184Z,yan2018spatial}. 

\subsection{Pedestrian Action Prediction}
Pedestrian action prediction formulates the prediction task as a binary classification problem where the objective is to determine if a pedestrian will start crossing in the near future as illustrated in Figure \ref{fig:presentation}. Being a sub-problem within action recognition, most of the existing approaches in the literature rely on the same modalities used for the latter. Preliminary works \cite{rasouli2017ICCVW,2018arXiv181009805V} formulated the problem as a static image classification problem to infer actions in a single image of a pedestrian. Afterward, approaches were designed to consider the temporal coherence in short-term motions of RGB images \cite{saleh2019real} and combined them with pose-based features \cite{2020arXiv200507796P,kotseruba2021benchmark,2015arXiv150308909Y}, increasing the size of their overall approaches drastically since both modalities needed to be extracted.
Some works rely on generative models to predict future actions representations which are then sent to a classifier \cite{chaabane2020looking,8794278}. All those methods present a drawback: they become sensitive to noise, background, and illumination conditions by including scene images in their approaches.
To overcome these issues, intention prediction only based on 2D body poses sequences has been explored with various available learning architectures such as convolutions \cite{fang2018pedestrian}, recurrent cells \cite{9136126,8500657}, graph-based models \cite{8917118} and proposed to enhance pose-based approaches by creating features based on body structure to capture different aspects of the data \cite{ranga2020vrunet,gesnouin2020predicting}. However, the lack of a common evaluation criterion, of normalized modalities inputs, of a common observation frames selection method, and common prediction horizons made the task of comparing each approach's robustness difficult if not impossible to realize. Lately, common evaluation protocols and modalities inputs \cite{kotseruba2021benchmark} were proposed to advance research on pedestrian action prediction further and obtain a fair comparison between all the upcoming methods.

\section{Methodology}
\begin{figure*}
    \centering
    \includegraphics[scale=0.157]{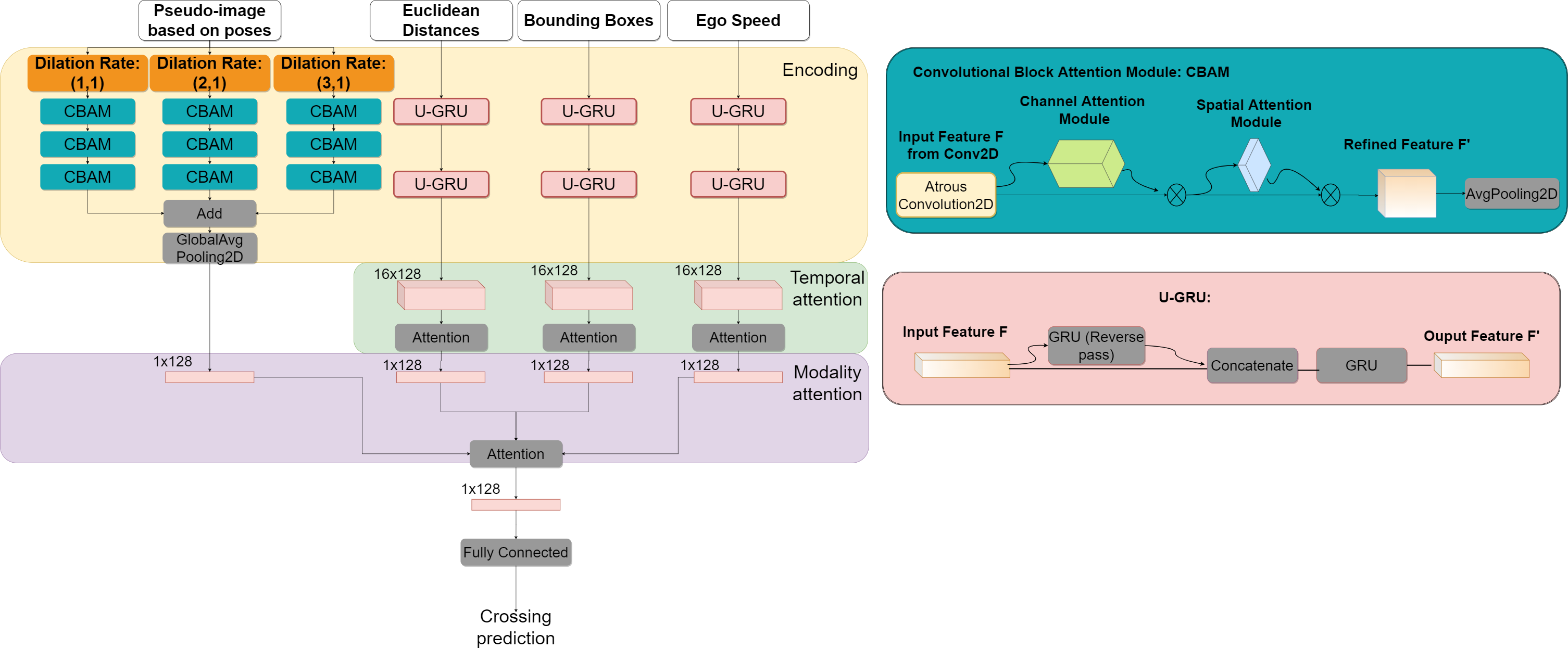}
    \caption{\textbf{The network architecture of TrouSPI-Net:} Its inputs consist in a sequence of 2D body poses transformed into a pseudo-image, relative pairwise distances of skeletal joints, bounding boxes, and ego-vehicle speed. U-GRUs encode every feature except pseudo-images, and each is fed into a temporal attention block. Pseudo-images are processed by parallel atrous CBAM \cite{woo2018cbam} blocks with different dilation rates and then added into a single vector in order to make the size of the pseudo-images block equal to the size of the U-GRUs outputs. Modality attention is then applied to the outputs of each branch, and the weighted outputs are fed into the fully connected layer. \textbf{U-GRU blocks:} the first GRU layer does the reverse pass, we then concatenate its output with the input data and finally compute the second GRU layer's output with a forward pass. \textbf{CBAM blocks:} given an intermediate feature map extracted by atrous 2D convolutions, the module sequentially infers attention maps along two separate dimensions: channel and spatial.}
    \label{fig:trouspi}
\end{figure*}

Based on the newly proposed evaluation procedures and inputs, we propose a new model for pedestrian action prediction based on 2D body poses: TrouSPI-Net, which is a largely modified and significantly improved version of the SPI-net architecture \cite{gesnouin2020predicting}. The diagram of the model is shown in Figure \ref{fig:trouspi} and the implementation details follow below.

\subsection{Extracting spatio-temporal features via parallel atrous convolutions on pseudo-images}

Pedestrian body poses sequences are defined as a vector: 

\begin{equation}
    \mathbf{s}=\left(\mathbf{s}_{1}, \mathbf{s}_{2}, \ldots, \mathbf{s}_{m}\right) \in \mathbf{R}^{m \times N \times d}
\end{equation}

where $m$ is the sequence duration, $N$ is the count of key-points, and $d$ is the dimension of each key-point. All sequences of skeletons are then sampled in the form of a 3-dimensional $(m, N, d)$-shaped tensor representing a 2D image-like spatio-temporal continuous representation of the sequence of poses. The horizontal axis of each pseudo-image represents the key-points axis while the vertical axis represents the time axis. $(x,y)$ dimensions of each key-point are then mapped to $RG(B)$ channels.

By using a 2D-convolution-ready representation format, we extract multi-scale spatio-temporal features using standard computer-vision methods such as atrous convolutions and enhance the feature extraction modules by using Convolutional Block Attention Module (CBAM) \cite{woo2018cbam} for self-attention mechanisms in each branch. Since sequences are represented as pseudo-images, CBAM blocks act as self-attention mechanisms for time and space conjointly.

Each of the pseudo-images is directly fed to three parallel branches. All three branches present a similar architecture designed for single-scale spatio-temporal feature extraction. In each branch, the pseudo-image is passed to an atrous CBAM block, illustrated in Figure \ref{fig:trouspi}, followed by a pooling layer. This process is repeated two more times. The difference between the three atrous CBAM blocks resides in the value of the dilation rate fixed in each branch. Having three different dilation rates for the spatio-temporal convolution layers allows the network to directly work at different time resolutions while staying at the same spatial resolutions. Moreover, compared to using different kernel sizes for each convolution, working with atrous convolution does not harm the model size. The outputs of the three branches extracting multi-scale spatio-temporal features are then summed into a single vector for later stages.

Formally, let $h^{(l, \beta)}(m,n)$ represent the input of the $l$-th atrous CBAM block of the $\beta$ branch, $K^{(l, \beta)}$ be the number of feature maps, $W_{k}^{(l, \beta)}(i,j)$ the $k$-th convolution filter of the $l$-th convolution in the $\beta$ branch with the length and the width of $m$ and $n$, $b_{k}^{(l, \beta)}$ the bias shared for the $k$-th filter map, $(r_{1}^{(l, \beta)},r_{2}^{(l, \beta)})$ the dilation rates and $\sigma$ an activation function. The intermediate feature map  $F(m,n)$ obtained by atrous 2D convolutions of the $l+1$-th CBAM block is calculated as:


\small
\begin{equation}
F(m, n)=\sigma\left(\sum_{k=1}^{K}\sum_{i=1}^{M} \sum_{j=1}^{N} h^{(l, \beta)}(m+r_{1}\times i, n+r_{2}\times j)\times W(i,j)+b\right)
\end{equation}
\normalsize

Where $K = K^{(l, \beta)}$, $(r_{1},r_{2}) = (r_{1}^{(l, \beta)},r_{2}^{(l, \beta)})$, $W = W_{k}^{(l, \beta)} $ and $b = b_{k}^{(l, \beta)}$. The ouput of the CBAM block $h^{(l+1, \beta)}(m,n)$ is then computed by sequentially inferring a 1D channel attention map $\mathrm{M}_{\mathrm{c}}$ and a 2D spatial attention map $\mathrm{M}_{\mathrm{s}}$ following the original recommendations of the CBAM paper \cite{woo2018cbam} and as illustrated in Figure \ref{fig:trouspi}:

\begin{equation}
\begin{array}{c}
\mathbf{F}^{\prime}(m,n)=\mathbf{M}_{\mathbf{c}}(\mathbf{F}(m,n)) \otimes \mathbf{F}(m,n) \\
h^{(l+1, \beta)}(m,n)=\mathbf{M}_{\mathbf{s}}\left(\mathbf{F}^{\prime}(m,n)\right) \otimes \mathbf{F}^{\prime}(m,n)
\end{array}
\end{equation}

where $\otimes$ denotes element-wise multiplication. Finally, the output $h^{(l+1, \beta)}(m,n)$ serves as the input of the batch normalization and pooling layer that directly follow the atrous CBAM block.

In our experiments, we have three branches: low resolution, medium resolution, high resolution branches $r_{1}^{(l, \beta)} \in [1 ; 3]$, $r_{2}^{(l, \beta)} = 1$,  $\beta \in [1 ; 3]$, three atrous CBAM blocks and pooling layers in each branch: $l \in [1 ; 3]$. $K^{(l, \beta)} = 64$ feature maps for each layer. Each convolution uses 3x3 kernels and is followed by a batch normalization layer. All the neurons use the LeakyRelu activation function: $\sigma(x)=max(0.2\mathrm{x},\mathrm{x})$, with the
exception of the $\mathrm{M}_{\mathrm{c}}$ and $\mathrm{M}_{\mathrm{s}}$ neurons which use the same hyper-parameters settings than the original CBAM paper.

\subsection{Modeling Location-viewpoint Invariant Features via U-GRUs}
To extract skeletal pose kinematic features invariant to locations and viewpoint, we represent a pose sequence as its evolution of skeletal joints relative Euclidean distances over time with the Joint Collection Distances (JCD) feature \cite{yang2019make}. The JCD feature is then flattened to a vector of dimension $m*\binom{N}{2}$.  Euclidean distances features are then processed with two U-GRUs blocks as illustrated in Figure \ref{fig:trouspi}.

Compared to regular Bidirectional GRUs, where the output layer can get information from past and future states simultaneously but are most sensitive to the input values around time $t$, in U-GRUs, past and future interact but in a limited way. U-GRUs allow the model to accumulate information while knowing which part of the information will be useful in the future and therefore exploit long-term temporal patterns on invariant locations and viewpoint skeletal dynamics. This compensates for the inabilities of the first pseudo-images stream to learn long-range temporal patterns and therefore acts as a regularizer for time and space. 

Similarly, context features such as bounding box positions and ego-vehicle speed are processed in parallel through the same U-GRUs architecture.

\begin{table*}[]
\caption{Evaluation results for baseline and state-of-the-art models and their variants on PIE and JAAD data-sets. Dashed lines separate different types of architectures. Modalities correspond to the type of networks used in the given approach, Model Params corresponds to the size of the network compiled on the benchmark \cite{kotseruba2021benchmark} with Additional Costs (Optical flow, Body Pose, RGB features) already extracted.
}
\resizebox{\textwidth}{!}{%
\label{table1}
\begin{tabular}{lll!{\vrule width \heavyrulewidth}lllll!{\vrule width \heavyrulewidth}lllll!{\vrule width \heavyrulewidth}lllll} 
\toprule
\multirow{2}{*}{ \textbf{Model Name } }       & \multirow{2}{*}{\textbf{Model Variants } } & \multicolumn{1}{l}{\multirow{2}{*}{\textbf{Model Params (Additional Costs) } }} & \multicolumn{5}{l}{\hspace{2.1cm}\textbf{$PIE$ } }                                                                 & \multicolumn{5}{l}{\hspace{1.8cm}\textbf{$JAAD_{behavior}$ } }                                                     & \multicolumn{5}{l}{\hspace{1.8cm}\textbf{$JAAD_{all}$} }                                          \\
                                              &                                            & \multicolumn{1}{l}{}                                                      & \textbf{ACC }  & \textbf{AUC }  & \textbf{F1 }   & \textbf{P }    & \multicolumn{1}{l}{\textbf{R } } & \textbf{ACC }  & \textbf{AUC }  & \textbf{F1 }   & \textbf{P }    & \multicolumn{1}{l}{\textbf{R } } & \textbf{ACC }  & \textbf{AUC }  & \textbf{F1 }   & \textbf{P }    & \textbf{R }     \\ 
\hline
\multirow{2}{*}{Static}                       & VGG16 \cite{2014arXiv1409.1556S}                                      & 14.7M                                                 & 0.71           & 0.60           & 0.41           & 0.49           & 0.36          & 0.59           & 0.52           & 0.71           & 0.63           & 0.82           & 0.82           & 0.75           & 0.55           & 0.49           & 0.63            \\
                                              & Resnet50 \cite{he2016deep}                                   & 23.6M                                                 & 0.70           & 0.59           & 0.38           & 0.47           & 0.32          & 0.46           & 0.45           & 0.54           & 0.58           & 0.51           & 0.81           & 0.72           & 0.52           & 0.47           & 0.56            \\
ATGC \cite{rasouli2017ICCVW}                                         & AlexNet                                    & 58.3M                                  & 0.59           & 0.55           & 0.39           & 0.33           & 0.47          & 0.48           & 0.41           & 0.62           & 0.58           & 0.66           & 0.67           & 0.62           & \textbf{0.76}  & \textbf{0.72}  & 0.80            \\ 
\hdashline
\multirow{2}{*}{ConvLSTM \cite{shi2015convolutional}}                     & VGG16                                      & 0.001M (VGG)                                          & 0.58           & 0.55           & 0.39           & 0.32           & 0.49          & 0.53           & 0.49           & 0.64           & 0.64           & 0.64           & 0.63           & 0.57           & 0.32           & 0.24           & 0.48            \\
                                              & ResNet50                                   & 0.001M (Resnet)                                          & 0.54           & 0.46           & 0.26           & 0.23           & 0.29          & 0.59           & 0.55           & 0.69           & 0.68           & 0.70           & 0.63           & 0.58           & 0.33           & 0.25           & 0.49            \\
SPI-Net \cite{gesnouin2020predicting}                                       & CNN  MLP                 & 0.1M (OpenPose)                                       & 0.66           & 0.54           & 0.30           & 0.35           & 0.27          & 0.58           & 0.55           & 0.66           & 0.67           & 0.65           & 0.81           & 0.72           & 0.52           & 0.48           & 0.58            \\
\multirow{2}{*}{SingleRNN \cite{9304591}}                    & LSTM                                       & 1.4M (2*VGG,OpenPose)                                   & 0.83           & 0.77           & 0.67           & 0.70           & 0.64          & 0.58           & 0.54           & 0.67           & 0.67           & 0.68           & 0.65           & 0.59           & 0.34           & 0.26           & 0.49            \\
                                              & GRU                                        & 1.0M (2*VGG,OpenPose)                                   & 0.81           & 0.75           & 0.64           & 0.67           & 0.61          & 0.51           & 0.48           & 0.61           & 0.63           & 0.59           & 0.78           & 0.75           & 0.54           & 0.44           & 0.70            \\
MultiRNN \cite{bhattacharyya2018long}                                     & GRU                                        & 1.8M (2*VGG,OpenPose)                                   & 0.83           & 0.80           & 0.71           & 0.69           & 0.73          & 0.61           & 0.50           & 0.74           & 0.64           & 0.86           & 0.79           & 0.79           & 0.58           & 0.45           & 0.79            \\
StackedRNN \cite{2015arXiv150308909Y}                                   & GRU                                        & 2.6M (2*VGG,OpenPose)                                   & 0.82           & 0.78           & 0.67           & 0.67           & 0.68          & 0.6            & \textbf{0.6}   & 0.66           & \textbf{0.73}  & 0.61           & 0.79           & 0.79           & 0.58           & 0.46           & 0.79            \\
HierarchicalRNN \cite{7298714}                              & GRU                                        & 3M (2*VGG,OpenPose)                                     & 0.82           & 0.77           & 0.67           & 0.68           & 0.66          & 0.53           & 0.5            & 0.63           & 0.64           & 0.61           & 0.80           & 0.79           & 0.59           & 0.47           & 0.79            \\
SFRNN \cite{Rasouli2019PedestrianAA}                                       & GRU                                        & 2.6M (2*VGG,OpenPose)                                  & 0.82           & 0.79           & 0.69           & 0.67           & 0.70          & 0.51           & 0.45           & 0.63           & 0.61           & 0.64           & 0.84           & 0.84           & 0.65           & 0.54           & \textbf{0.84}   \\ 
\hdashline
C3D \cite{2014arXiv1412.0767T}                                          & RGB                                        & 78M                                              & 0.77           & 0.67           & 0.52           & 0.63           & 0.44          & 0.61           & 0.51           & 0.75           & 0.63           & \textbf{0.91}  & 0.84           & 0.81           & 0.65           & 0.57           & 0.75            \\
\multirow{2}{*}{I3D \cite{8099985}}                          & RGB                                        & 12.3M                                            & 0.80           & 0.73           & 0.62           & 0.67           & 0.58          & 0.62           & 0.56           & 0.73           & 0.68           & 0.79           & 0.81           & 0.74           & 0.63           & 0.66           & 0.61            \\
                                              & Optical flow                               & 12.3M (FlowNet2)                                     & 0.81           & 0.83           & 0.72           & 0.60           & \textbf{0.9}  & 0.62           & 0.51           & 0.75           & 0.65           & 0.88           & 0.84           & 0.80           & 0.63           & 0.55           & 0.73            \\
TwoStream \cite{NIPS2014_00ec53c4}                                     & VGG16                                      & 134.3M (FlowNet2)                               & 0.64           & 0.54           & 0.32           & 0.33           & 0.31          & 0.56           & 0.52           & 0.66           & 0.66           & 0.66           & 0.60           & 0.69           & 0.43           & 0.29           & 0.83            \\
PCPA \cite{kotseruba2021benchmark}                                         & Temp. +mod. attention                      & 31.2M (C3D,OpenPose)                                  & 0.87           & 0.86           & 0.77           & -              & -             & 0.58           & 0.5            & 0.71           & -              & -              & \textbf{0.85}  & \textbf{0.86}  & 0.68           & -              & -               \\ 
\hdashline
\multirow{2}{*}{\textbf{TrouSPI-Net} (ours) } & CBAM attention block                       & 1.5M\textasciitilde{} (OpenPose)                      & \textbf{0.88}  & \textbf{0.88}  & \textbf{0.80}  & 0.73           & 0.89          & \textbf{0.64}  & 0.56           & \textbf{0.76}  & 0.66           & \textbf{0.91}  & \textbf{0.85}  & 0.73           & 0.56           & 0.57           & 0.55            \\
                                              & SE attention block                         & 1.5M (OpenPose)                                       & \textbf{0.88}  & 0.87           & \textbf{0.80}  & \textbf{0.77}  & 0.84          & \textbf{0.64}  & 0.55           & \textbf{0.76}  & 0.65           & \textbf{0.91}  & 0.82           & 0.77           & 0.58           & 0.49           & 0.70            \\
\bottomrule
\end{tabular}}
\end{table*}

\subsection{Combining all the features branches}
Following the successful application of temporal attention and modality attention in multi-modal approaches for pedestrian action prediction, we finally apply the same temporal attention and modality attention mechanisms used in PCPA \cite{kotseruba2021benchmark} to all our features branches to fuse them effectively.
Nonetheless, the nature of the inputs merged in TrouSPI-net is entirely different compared to the initial multi-modal PCPA \cite{kotseruba2021benchmark} architecture. While PCPA \cite{kotseruba2021benchmark} merges inputs such as sequences of RGB camera images processed by 3D convolution and poses processed via simple recurrent networks without spatio-temporal coherence of body actions, TrouSPI-Net was designed to operate without needing additional RGB scene-context and uses different body poses representations that were encoded to treat the spatial and the temporal information of body action for different time resolutions.

For each feature extracted by U-GRUs: we apply temporal attention \cite{kotseruba2021benchmark} to weight the relative importance of frames in the observation relative to the last seen frame. We then apply modality attention \cite{kotseruba2021benchmark} to the weighted outputs of the U-GRUs features and the output of the pseudo-images stream. This fuses inputs from multiple modalities into a single representation by weighted summation of the information from individual modalities. The output of the modality attention block is finally passed to a dense layer for prediction.

\section{Experiments}
To evaluate the presented multi-branch approach and several variations of its architecture, we conducted experiments on two large public data-sets for studying pedestrian behaviors in traffic: JAAD \cite{rasouli2017ICCVW,Rasouli2017IV} and PIE \cite{rasouli2017they}. JAAD contains 346 clips and focuses on pedestrians intending to cross, PIE contains 6 hours of continuous footage and provides annotations for all pedestrians sufficiently close to the road regardless of their intent to cross in front of the ego-vehicle and provides more diverse behaviors of pedestrians.

\subsection{Evaluation Setup}
We base our experiments on the newly proposed evaluation criteria \cite{kotseruba2021benchmark} with common evaluation protocols, splits and normalized modalities inputs. As provided in the new benchmark, observation data for each pedestrian is sampled so that the last frame of observation is between 1s and 2s before the crossing event. We report the results using regular classification metrics: accuracy, AUC, precision, recall and $F_{1}$-score given by $F_{1}=\frac{2 \times \text { precision } \times \text { recall }}{\text { precision }+\text { recall }}$.

In architecture variations and branch ablations studies, we explore how each TrouSPI-Net component contributes to the pedestrian action prediction performance by removing one component while keeping others unchanged. We also explore the performance of CBAM blocks in the pseudo-image stream by comparing them to similar self-attention blocks designed for 2D convolutions: Squeeze and
Excitation method (SE blocks) \cite{hu2018squeeze}. Finally, we explore the impact of adding a second modality to TrouSPI-Net by using 3D convolutions \cite{2014arXiv1412.0767T} on the local box feature available in the data-set.

\begin{table}[]
\caption{Architecture variations and Ablation studies for TrouSPI-Net on PIE data-set.
}
\label{table2}
\resizebox{8.5cm}{!}{%
\centering
\begin{tabular}{lllll} 
\toprule
\textbf{Model Variants (Additional Costs)}                               & \textbf{Params} & \textbf{ACC}  & \textbf{AUC}  & \textbf{F1}    \\ 
\hline
TrouSPI-Net without euclidean distances      & 1.4M            & 0.87           & 0.85           & 0.78            \\
TrouSPI-Net without parallel atrous branches & 0.8M            & 0.86           & 0.80           & 0.72            \\ 
\hdashline
{TrouSPI-Net GRUs}                     & 1.3M            & 0.85           & 0.80           & 0.72            \\
TrouSPI-Net BiGRUs                           & 1.6M            & 0.86           & 0.82           & 0.75            \\ 
\hdashline
TrouSPI-Net without attention Block          & 1.4M            & 0.87           & 0.85           & 0.78            \\
TrouSPI-Net with SE attention Block          & 1.5M            & \textbf{0.88}  & 0.87           & \textbf{0.80}   \\ 
\hdashline
\textbf{TrouSPI-Net }                                 & 1.5M            & \textbf{0.88}  & \textbf{0.88}  & \textbf{0.80}   \\
TrouSPI-Net with two modalities (C3D) & 30.2M           & \textbf{0.88}           & 0.87           & \textbf{0.80}            \\
\bottomrule
\end{tabular}}
\end{table}

\subsection{Implementation Details}
We use U-GRUs with 64 hidden units for encoding all features, except the pseudo-image. L2 regularization of 0.001 is added to the final dense layer and a dropout of 0.5 is added after the attention block. The number of observation frames \textit{m} is set to 16. Body poses extracted by OpenPose \cite{cao2017realtime} and proposed in the benchmark \cite{kotseruba2021benchmark} are sampled in the form of a 3-dimensional (16,18,2)-shaped tensor for the pseudo-images stream and 2-dimensional (16,153)-shaped tensor for the U-GRUs stream. The ego-vehicle speed feature is used only in the PIE data-set and omitted in JAAD. To compensate for the significant class imbalance, we apply class weights inversely proportional to the percentage of samples of each class in each data-sets.
We train the model with Ranger Optimizer: a combination of Lookahead ($k=6, \alpha=0.5$) \cite{2019arXiv190708610Z} and Radam \cite{2019arXiv190803265L}, binary cross-entropy loss and batch size set to 8. We train for 80 epochs with learning rate set to 5.0e-05 for PIE and 5.0e-06 for JAAD.

\subsection{Discussion}
The results of the final TrouSPI-Net model are presented in Table \ref{table1}. Results are most improved compared to State-of-the-Art on the PIE data-set, where accuracy is increased by 1\%, AUC by 2\% and $F_{1}$-score by 3\% compared to PCPA \cite{kotseruba2021benchmark}, a model with two perception modalities: RGB images and poses. On JAAD, our model performs comparably if not better with state-of-the-art across some metrics. This leads us to believe that approaches using only one additional network to compute perception modalities can be competitive with approaches that combine multiple. 

A comparison of $F_{1}$-scores between our approach and the best-performing methods that exist at this day shows that our approach offers better $F_{1}$-scores for two out of three benchmarks. It shows that TrouSPI-Net is more balanced than other approaches for the task of pedestrian crossing prediction. Finally, results obtained by TrouSPI-Net on $JAAD_{all}$ should be taken with a pinch of salt since the data-set considers all the visible pedestrians who are far away from the road and are not crossing. Since pose estimation algorithms are still struggling with scale to extract informative poses for people at the back of a scene, TrouSPI-Net does not manage to extract discriminating features because of the low quality of the poses extracted and relies mainly on other features to realize its inference. This explains its lower performance compared to the two other benchmarks. However, it should not be considered as an issue since those pedestrians are not directly interacting with the vehicle in any way. If they were to become a danger in the future, they would have to step closer to it, and therefore pose estimation algorithms should be able to extract informative poses.

\subsubsection{\textbf{Architecture variations and branch ablations}}
Removing the parallel atrous branches from the pseudo-image stream leads to a degradation of the performance indicators (Acc, AUC, $F_{1}$) on PIE data-set by respectively, 2\%, 8\% and 8\%. Similarly, removing the stream acting as a regularizer with relative distances degrades the performance indicators by respectively 1\%, 3\%  and 2\%. Therefore, we can highlight the importance of the three parallel branches to extract spatio-temporal features for different time scales and the importance of the euclidean distances stream to act as a regularizer for the overall approach performance.

Secondly, we evaluate the importance of using a spatio-temporal attention module over the parallel pseudo-images extraction module. We first disregard spatio-temporal attention completely in the given pseudo-images stream and then replace CBAM blocks \cite{woo2018cbam} with SE blocks \cite{hu2018squeeze}. Experimental results show that removing the attention-enhanced 2D atrous convolutions degrades the performance indicators by respectively 1\%, 3\%, 2\%, whereas replacing CBAM blocks \cite{woo2018cbam} by SE blocks \cite{hu2018squeeze} do not drastically impact TrouSPI-Net's performance and even increases it across some metrics according to Table \ref{table1}. In conclusion, introducing a spatio-temporal attention module over the parallel features extraction module seems to improve our model performance. Future studies could fruitfully explore this further by introducing a custom spatio-temporal attention module specifically designed for the parallel pseudo-images extraction module.

Finally, we evaluate the importance of U-GRUs by replacing them with GRUs and Bidirectional GRUs. Table \ref{table2} results show that both modified approaches lead to a degradation of the performance indicators by respectively 3\%, 8\%, 8\% and 2\%, 6\%, 5\%. Therefore, we can highlight the importance of U-GRUs to exploit the bidirectional temporal and long-term contexts compared to other state-of-the-art approaches designed to capture sequential features. It also leads us to believe that an effective pedestrian action prediction model should focus on both long-term dependencies and multi-scale short temporal features to be effective. 

\begin{table}[!h]
\centering
\caption{Architecture comparison of floating point operations per second (FLOPS) in millions, Cuda Memory Usage (CMU) in Megabytes and Weights Memory Requirements (WMR) in Megabytes. RGB features extracted by CNNs are taken in consideration during computations.}
\label{table3}
\resizebox{\columnwidth}{!}{%
\begin{tabular}{l!{\vrule width \lightrulewidth}l!{\vrule width \lightrulewidth}l!{\vrule width \lightrulewidth}l} 
\toprule
\multicolumn{1}{l}{\textbf{Model(Additional Costs)}} & \multicolumn{1}{l}{\textbf{FLOPS (Mio.)}} & \multicolumn{1}{l}{\textbf{CMU (MB)}} & \textbf{WMR (MB)}  \\ 
\midrule
VGG16 \cite{2014arXiv1409.1556S}                                  & 29.4                                      & 72.1                                  & 56.1               \\
Resnet50 \cite{he2016deep}                               & 47.0                                      & 47.0                                  & 90.0               \\ 
\hdashline
ConvLSTM \cite{shi2015convolutional} (VGG)                         & 29.5                                      & 93.5                                  & 56.2               \\
SingleRNN \cite{9304591} (2 VGG)      & 65.3                                      & 145.3                                 & 60.0               \\
MultiRNN \cite{bhattacharyya2018long} (2 VGG)       & 71.6                                      & 146.0                                 & 63.0               \\
StackedRNN \cite{2015arXiv150308909Y} (2 VGG)     & 76.3                                      & 146.8                                 & 66.0               \\
SFRNN \cite{Rasouli2019PedestrianAA} (2 VGG)          & 73.6                                      & 146.5                                 & 64.5               \\ 
\hdashline
C3D \cite{2014arXiv1412.0767T}                                    & 156.0                                     & 182.6                                 & 297.5              \\
I3D \cite{8099985}                                    & 24.6                                      & 334.1                                 & 46.9               \\
PCPA \cite{kotseruba2021benchmark} (C3D)                             & 220                                       & 320.2                                 & 414.9              \\ 
\hdashline
SPI-net \cite{gesnouin2020predicting}                               & \textbf{0.3}                                       & \textbf{2.5}                                   & \textbf{0.3}                \\
\textbf{TrouSPI-Net} (ours)            & 3.0                                       & 6.8                                   & 5.4                \\
TrouSPI-Net with two modalities (C3D)            & 216.7                                       & 322.6                                   & 412.9                \\
\bottomrule
\end{tabular}}
\end{table}

\subsubsection{\textbf{Using a second perception modality with TrouSPI-Net}}
One of the main advantages of using a scene-agnostic model using such sparse perception modality instead of aggregating multiple perception modalities is the smaller model size leading to an easier deployment into embedded devices, as table \ref{table3} shows.
Moreover, when combined with 3D convolutions of cropped images, including the pedestrians, TrouSPI-Net's computational costs grown dramatically without gaining any performance on PIE data-set as tables \ref{table2} and \ref{table3} show. 
 This may be considered a further validation of pose-based only networks for Pedestrian Action Prediction as lightweight models designed for embedded devices with real-time constraints, which do not need additional context input to work effectively. 
 While this affirmation is established for pedestrians where pose estimation inferences are possible and with limited occlusions, the question remains open for scenes with very high occlusions between pedestrians, occlusions between pedestrians and scene objects, or abnormal behaviors such as crowd movement. For those cases, implementing a way to treat Static RGB images effectively as a context feature might still prove important.

\subsubsection{\textbf{The drawbacks of relying on additional networks to extract perception modalities}}
Although other models also apply additional networks to extract multiple perception modalities such as pose, flow or background context and the proposed approach beats the state-of-the-art while being smaller in comparison according to Tables \ref{table1} and \ref{table3}, its application also relies on one additional algorithm to operate. If TrouSPI-Net was to be implemented outside of the JAAD and PIE benchmark, one would have to add to TrouSPI-Net's size the pose extraction model used to compute the pose information. In our case, OpenPose \cite{cao2017realtime} was used to compute the inputs available in \cite{kotseruba2021benchmark}. Therefore, the overall approach is $\sim$ 53.5M parameters. However, it leads to a practical methodology as interchanging the additional approaches to extract poses does not jeopardize the TrouSPI-Net approach. Contrary to image-based approaches, if improvements such as inference time or average precision by key-points were made in the field of pose estimation, TrouSPI-Net could still be applicable without any modification. 

Moreover, the proposed benchmark \cite{kotseruba2021benchmark} currently omits a major issue for pedestrian intention prediction: temporal tracking of pedestrians to avoid mixing identities over time. Such questions are rarely raised and approaches mainly rely on the ground-truth IDs of each pedestrian. However, such concerns are mandatory to easily transpose the pedestrian action prediction approaches into real-life scenarios without pedestrians' ground-truth IDs. 

In TrouSPI-Net's case, to ensure a better follow-up of the protagonists in the scene and avoid mixing the identities of two protagonists, one could for example replace OpenPose \cite{cao2017realtime} by pose estimation networks sequentially based on pose matching for tracking \cite{2018arXiv180200977X,2019arXiv190502822N,Raaj_2019_CVPR}. Such a substitution would provide the TrouSPI-Net model every modality it needs to work in a non-controlled environment with only one additional network: body poses, handcrafted body poses features, bounding boxes positions of the pedestrians and their respective individual ID's.

\addtolength{\textheight}{3cm}   

\section{CONCLUSIONS}
We introduced a new lightweight multi-branch neural network to predict pedestrians' actions using only one additional network to extract perception modalities: 2D pedestrian body poses. The proposed TrouSPI-Net model largely extends and improves the SPI-Net \cite{gesnouin2020predicting} approach in several ways. First, we introduce parallel processing branches to allow the architecture to access different time resolutions with atrous convolutions enhanced with self-attention mechanisms. Secondly, we apply U-GRUs on the evolution of relative Euclidean body distances over time, which acts as a regularizer of the first stream for both time and space. We then extend the U-GRUs approach as one baseline method to consider long-term temporal coherence and process each sequence of context features such as bounding box positions or ego-vehicle speed with U-GRUs. Finally, following the newly proposed evaluation procedures and benchmarks for JAAD and PIE (two challenging pedestrian action prediction data-sets), our experimental results show that TrouSPI-Net achieved 76\% F1 score on JAAD and 80\% F1 score on PIE, therefore outperforming current state-of-the-art. This shows that using only body poses can outperform approaches that combine multiple networks to extract different perception modalities. Subsequently, our model inherits interesting properties such as being completely invariant to any scene-background context, leading to a lightweight approach focusing only on the pedestrian's movement. Therefore, we believe that TrouSPI-Net could be an interesting baseline to easily compare to for future works aiming at developing a pose-only based model for pedestrian intention prediction and has the potential to improve many other human action recognition or prediction tasks.

\section{ACKNOWLEDGEMENT}
The authors acknowledge the infrastructure and support of the PELOPS unit and the interdisciplinary R\&D department of the Vedecom institute. The authors would also like to thank Ahmet Erdem, Thomas Gilles and Raphaël Rozenberg for  the helpful discussions regarding U-GRUS, Marie Morel and Emmanuel Doucet for their fruitful comments and corrections on the manuscript.


{\small
\bibliographystyle{ieee}
\bibliography{egbib}
}

\end{document}